\documentclass[10pt,twocolumn,letterpaper]{article}

\usepackage[pagenumbers]{cvpr} 

\usepackage[dvipsnames]{xcolor}

\usepackage{graphicx}
\usepackage{float}
\usepackage{multirow}
\usepackage{makecell}
\usepackage{pifont}
\usepackage{listings}
\usepackage{enumitem}
\usepackage{color, colortbl}
\usepackage{wrapfig}

\newcolumntype{x}[1]{>{\centering\arraybackslash}p{#1pt}}
\newcolumntype{y}[1]{>{\raggedright\arraybackslash}p{#1pt}}
\newcolumntype{z}[1]{>{\raggedleft\arraybackslash}p{#1pt}}
\newcommand{\tablestyle}[2]{\setlength{\tabcolsep}{#1}\renewcommand{\arraystretch}{#2}\centering\footnotesize}

\definecolor{mgreen}{RGB}{1,150,74}
\newcommand\up[1]{\textcolor{mgreen}{$^{\uparrow{#1}}$}}
\newcommand\down[1]{\textcolor{red}{$^{\downarrow{#1}}$}}

\definecolor{mblue}{HTML}{6895D2}
\definecolor{morange}{HTML}{E48F45}
\definecolor{mpurple}{HTML}{836096}
\definecolor{mpink}{HTML}{CD6688}

\newcommand\blfootnote[1]{%
\begingroup
\renewcommand\thefootnote{}\footnote{#1}%
\addtocounter{footnote}{-1}%
\endgroup
}
\usepackage[accsupp]{axessibility}

\definecolor{cvprblue}{rgb}{0.21,0.49,0.74}
\usepackage[pagebackref,breaklinks,colorlinks,citecolor=cvprblue]{hyperref}

\def\paperID{9127} 
\def\confName{CVPR}
\def\confYear{2024}

\title{UniPAD: A Universal Pre-training Paradigm for Autonomous Driving}

\author{Honghui Yang$^{1,2*}$,
Sha Zhang$^{2,6}$, Di Huang$^{2,7}$, Xiaoyang Wu$^{2,5}$, Haoyi Zhu$^{2,6}$, Tong He$^{2\dag}$\\
Shixiang Tang$^2$, Hengshuang Zhao$^5$, Qibo Qiu$^{8}$, Binbin Lin$^{3,4\dag}$, Xiaofei He$^1$, Wanli Ouyang$^2$\\
$^1$State Key Lab of CAD\&CG, Zhejiang University~~
$^2$Shanghai Artificial Intelligence Laboratory\\
$^3$School of Software Technology, Zhejiang University~~
$^4$Fullong Inc.~~
$^5$HongKong University\\
$^6$University of Science and Technology of China~~
$^7$The University of Sydney~~
$^8$Zhejiang Lab~~
}

\begin{document}
\maketitle
\begin{abstract}
In the context of autonomous driving, the significance of effective feature learning is widely acknowledged.
While conventional 3D self-supervised pre-training methods have shown widespread success, most methods follow the ideas originally designed for 2D images.
In this paper, we present UniPAD, a novel self-supervised learning paradigm applying 3D volumetric differentiable rendering. UniPAD implicitly encodes 3D space, facilitating the reconstruction of continuous 3D shape structures and the intricate appearance characteristics of their 2D projections. The flexibility of our method enables seamless integration into both 2D and 3D frameworks, enabling a more holistic comprehension of the scenes.
We manifest the feasibility and effectiveness of UniPAD by conducting extensive experiments on various 3D perception tasks.
Our method significantly improves lidar-, camera-, and lidar-camera-based baseline by 9.1, 7.7, and 6.9 NDS, respectively.
Notably, our pre-training pipeline achieves 73.2 NDS for 3D object detection and 79.4 mIoU for 3D semantic segmentation on the nuScenes validation set, achieving state-of-the-art results in comparison with previous methods.
\blfootnote{$^*$This work was done during his internship at Shanghai Artificial Intelligence Laboratory.}
\blfootnote{$^\dag$Corresponding author}
\end{abstract}
\section{Introduction}

Self-supervised learning for 3D point cloud data is of great significance as it is able to use vast amounts of unlabeled data efficiently, enhancing their utility for various downstream tasks like 3D object detection~\citep{yang2019std,deng2021voxelrcnn,shi2021pvrcnnplusplus,shi2020pvrcnn,yang2022graphrcnn,liu2023petrv2} and semantic segmentation~\citep{ChoyGS2019mink,liu20223dqueryis,wu2022pointtransv2,zhong2023understanding,liu2023mars3d,liu2023uniseg}.
Although significant advances have been made in self-supervised learning for 2D images~\citep{he2022mae,he2020moco,chen2021simsiam,chen2020simclr}, extending these approaches to 3D point clouds have presented considerably more significant challenges. This is partly caused by the inherent sparsity of the data, and the variability in point distribution due to sensor placement and occlusions by other scene elements.
Previous pre-training paradigms for 3D scene understanding adapted the idea from the 2D image domain and can be roughly categorized into two groups: contrastive-based and MAE-based. 

Contrastive-based methods~\citep{zhang2021depthcontrast,chen20224dconstrast} explore pulling similar 3D points closer together while pushing dissimilar points apart in feature space through a contrastive loss function. For example, PointContrast~\citep{xie2020pointcontrast} directly operates on each point and has demonstrated its effectiveness on various downstream tasks. 
Nonetheless, the sensitivity to positive/negative sample selection and the associated increased latency often impose constraints on the practical applications of these approaches. 
Masked AutoEncoding (MAE)~\citep{he2022mae}, which encourages the model to learn a holistic understanding of the input beyond low-level statistics, has been widely applied in the autonomous driving field.
Yet, such a pretext task has its challenges in 3D point clouds due to the inherent irregularity and sparsity of the data.
VoxelMAE~\citep{hess2022voxelmae} proposed to divide irregular points into discrete voxels and predict the masked 3D structure using voxel-wise supervision.
The coarse supervision may lead to insufficient representation capability.

\begin{figure}[!t]
    \centering
    \includegraphics[width=0.8\columnwidth]{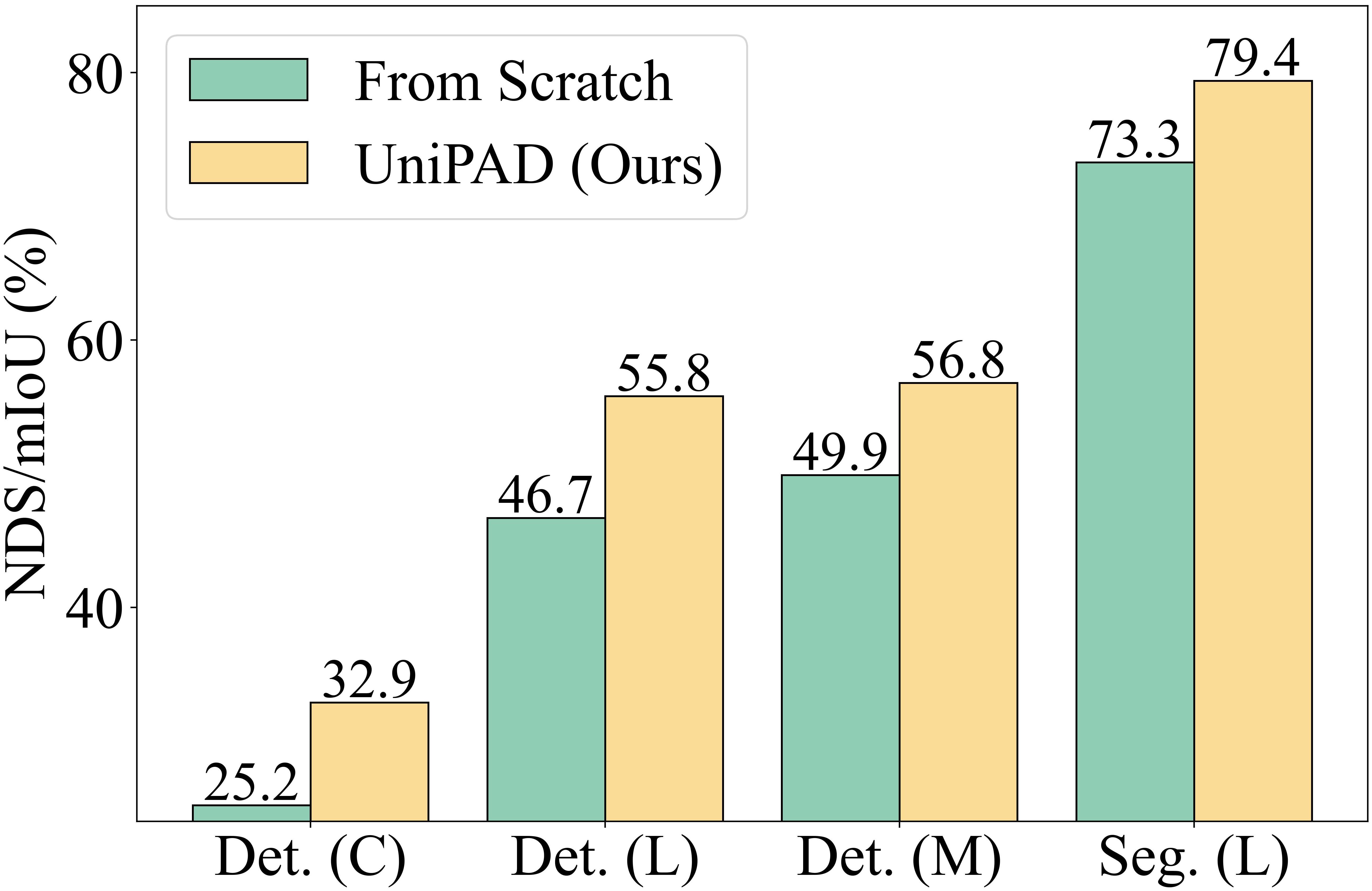}
    \vspace{-1.0em}
    \caption{Effect of our pre-training for 3D detection and segmentation on the nuScenes~\cite{caesar2020nuscenes} dataset, where C, L, and M denote camera, LiDAR, and fusion modality, respectively.}
    \label{fig:improve}
    \vspace{-1.8em}
\end{figure}

\begin{figure*}[!t]
	\centering
	\includegraphics[width=1.9\columnwidth]{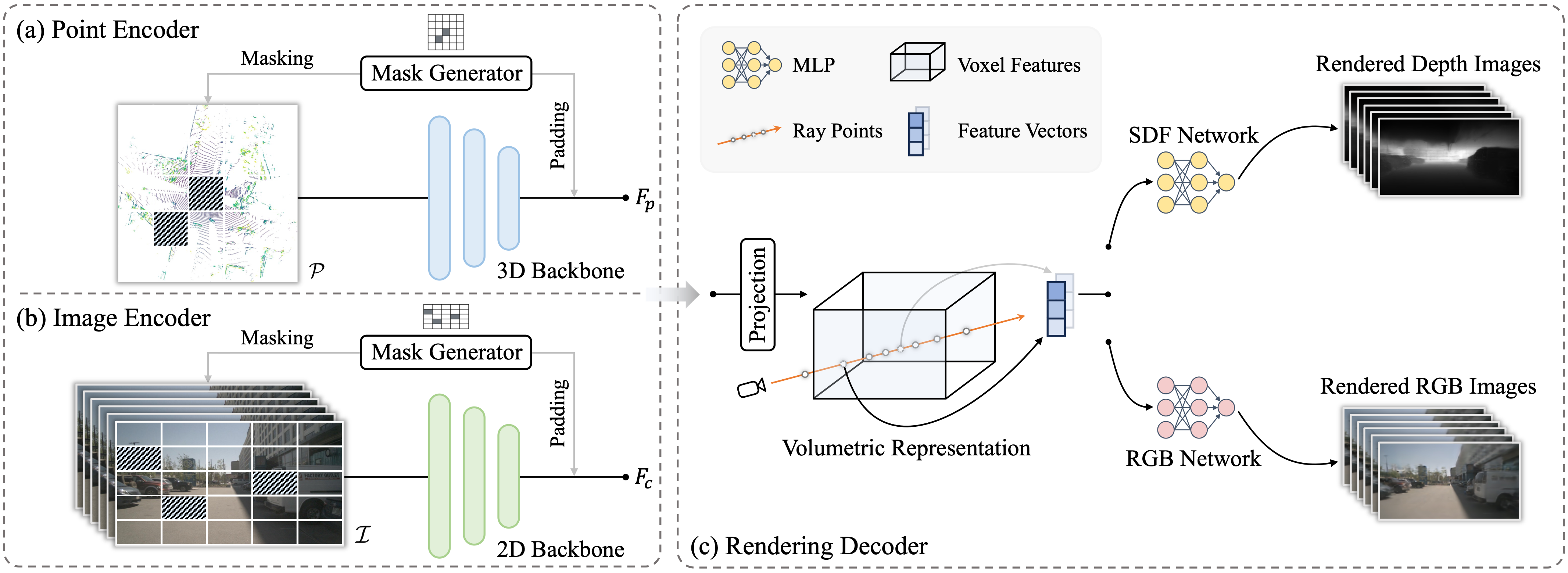}
	\vspace{-0.8em}
	\caption{The overall architecture.
	Our framework takes LiDAR point clouds or multi-view images as input.
	We first propose the mask generator to partially mask the input.
	Next, the modal-specific encoder is adapted to extract sparse visible features, which are then converted to dense features with masked regions padded as zeros.
	The modality-specific features are subsequently transformed into the voxel space, followed by a projection layer to enhance voxel features.
	Finally, volume-based neural rendering produces RGB or depth prediction for both visible and masked regions.}
	\label{fig:overall_stru}
	\vspace{-1.4em}
\end{figure*}

In this paper, we come up with a novel pre-training paradigm tailored for effective 3D representation learning, which not only eliminates the need for complex positive/negative sample assignments but also implicitly provides continuous supervision signals to learn 3D shape structures.
The whole framework, as illustrated in Figure~\ref{fig:overall_stru}, takes the masked point cloud as input and aims to reconstruct the missing geometry on the projected 2D depth image via 3D differentiable neural rendering. 

Specifically, when provided with a masked LiDAR point cloud, our approach employs a 3D encoder to extract hierarchical features. Then, the 3D features are transformed into the voxel space via voxelization. We further apply a differentiable volumetric rendering method to reconstruct the complete geometric representation.
The flexibility of our approach facilitates its seamless integration for pre-training 2D backbones. Multi-view image features construct the 3D volume via lift-split-shoot (LSS)~\citep{phi2020lss}.
To maintain efficiency during the training phase, we propose a memory-efficient ray sampling strategy designed specifically for autonomous driving applications, which can greatly reduce training costs and memory consumption. Compared with the conventional methods, the novel sampling strategy boosts the accuracy significantly.

Extensive experiments conducted on the competitive nuScenes~\citep{caesar2020nuscenes} dataset demonstrate the superiority and generalization of the proposed method.
For pre-training on the 3D backbone, our method yields significant improvements over the baseline, as shown in Figure~\ref{fig:improve}, achieving enhancements of \textbf{9.1} NDS for 3D object detection and \textbf{6.1} mIoU for 3D semantic segmentation, surpassing the performance of both contrastive- and MAE-based methods. Notably, our method achieves the state-of-the-art mIoU of \textbf{79.4} for segmentation on nuScenes dataset.
Furthermore, our pre-training framework can be seamlessly applied to 2D image backbones, resulting in a remarkable improvement of \textbf{7.7} NDS for multi-view camera-based 3D detectors.
We directly utilize the pre-trained 2D and 3D backbones to a multi-modal framework.
Our method achieves \textbf{73.2} NDS for detection, reaching the level of existing state-of-the-art methods.
Our contributions can be summarized as follows:
\vspace{-2pt}
\begin{itemize}[leftmargin=20pt,topsep=4pt,itemsep=2pt]
\item To the best of our knowledge, we are the first to explore the 3D differentiable rendering for self-supervised learning in the context of autonomous driving.
\item The flexibility of the method makes it easy to be extended to pre-train a 2D backbone. With a novel sampling strategy, our approach exhibits superiority in both effectiveness and efficiency.
\item We conduct comprehensive experiments on the nuScenes dataset, wherein our method surpasses the performance of six pre-training strategies. Experimentation involving seven backbones and two perception tasks provides convincing evidence for the effectiveness of our approach.
\end{itemize}

\section{Related Work}
\paragraph{Self-supervised learning in point clouds} has gained remarkable progress in recent years~\citep{chen20224dconstrast,li2022dpco,liang2021gcc3d,liu2022maskpoint,pang2022pointmae,tian2023geomae,xv2023mvjar,yin2022proposalcontrast,zhang2021depthcontrast,huang2023ponder,zhu2022x,min2022voxelmae,wu2023ppt,hess2022voxelmae}.
PointContrast~\citep{xie2020pointcontrast} contrasts point-level features from two transformed views to learn discriminative 3D representations.
Point-BERT~\citep{yu2022pointbert} introduces a BERT-style pre-training strategy with standard transformer networks.
OcCo~\cite{wang2021occo} occludes point clouds based on different viewpoints and learns to complete them.
PointContrast~\cite{xie2020pointcontrast} contrasts point-level features from two transformed views to learn discriminative 3D representations.
MSC~\citep{wu2023msc} incorporates a mask point modeling strategy into a contrastive learning framework.
PointM2AE~\citep{zhang2022pointm2ae} utilizes a multiscale strategy to capture both high-level semantic and fine-grained details.
STRL~\citep{huang2021spatio} explores the rich spatial-temporal cues to learn invariant representation in point clouds.
GD-MAE~\citep{yang2023gd-mae} applies a generative decoder for hierarchical MAE-style pre-training.
ALSO~\citep{boulch2023also} regards the surface reconstruction as the pretext task for representation learning.
Unlike previous works primarily designed for point clouds, our pre-training framework is applicable to both image-based and point-based models.

\vspace{-0.8em}
\paragraph{Representation learning in image} has been well-developed~\citep{bachmann2022multimae, bao2021beit, tong2022videomae, chen2023survey, wangw2023survey, wangx2023survey}, and has shown its capabilities in all kinds of downstream tasks as the backbone initialization.
Contrastive-based methods, such as MoCo~\citep{he2020moco} and MoCov2~\citep{chen2020improved}, learn the representations of images by discriminating the similarities between different augmented samples.
MAE-based methods~\citep{gao2022convmae, tian2023spark} obtain the promising generalization ability by recovering masked patches.
In autonomous driving, models pre-trained on ImageNet~\citep{deng2009imagenet} are widely utilized in image-related tasks~\citep{liu2022petr, liu2023sparsebev, liang2022bevfusion, li2022uvtr, yan2023cmt, hu2023uniad, li2022bevformer}.
For example, to compensate for the insufficiency of 3D priors in tasks like 3D object detection, depth estimation~\citep{park2021dd3d} and monocular 3D detection~\citep{wang2021fcos3d} are usually exploited as the additional pre-training techniques.

\vspace{-0.8em}
\paragraph{Neural rendering for autonomous driving} utilizes neural networks to differentially render images from 3D scene representation~\citep{chen2022tensorf,ben2020nerf,oechsle2021unisurf,xu2023nerfdet,xv2022pointnerf,yang2023unisim}.
Those methods can be roughly divided into two categories: perception and simulation.
Being capable of capturing semantic and accurate geometry, NeRFs are gradually utilized to do different perception tasks including panoptic segmentation~\citep{fu2022panoptic},  object detection~\citep{xu2023nerfdet, xu2023mononerd}, segmentation~\citep{kundu2022panoptic}, and instance segmentation~\citep{zhi2021place}.
For simulation, 
MARS~\citep{wu2023mars} models the foreground objects and background environments separately based on NeRF, making it flexible for scene controlling in autonomous driving simulation.
Considering the limited labeled LiDAR point clouds data, NeRF-LiDAR~\citep{zhang2023nerf_lidar} proposes to generate realistic point clouds along with semantic labels for the LiDAR simulation. 
Besides, READ~\citep{li2023read} explores multiple sampling strategies to make it possible to synthesize large-scale driving scenarios.
Inspired by them, we make novel use of NeRF, with the purpose of universal pre-training, rather than of novel view synthesis.

\section{Methodology}
The UniPAD framework is a universal pre-training paradigm that can be easily adapted to different modalities, e.g., 3D LiDAR point and multi-view images. 
Our framework is shown in Figure~\ref{fig:overall_stru}, which contains two parts, i.e., a modality-specific encoder and a volumetric rendering decoder. 
For processing point cloud data, we employ a 3D backbone for feature extraction. In the case of multi-view image data, we leverage a 2D backbone to extract image features, which are then mapped into 3D space to form the voxel representation.
Similar to MAE~\cite{he2022mae}, a masking strategy is applied for the input data to learn effective representation.
For decoders, we propose to leverage off-the-shelf neural rendering with a well-designed memory-efficient ray sampling. 
By minimizing the discrepancy between rendered 2D projections and the input, our approach encourages the model to learn a continuous representation of the geometric or appearance characteristics of the input data. 

\subsection{Modal-specific Encoder}
\label{subsec:sparse_encoder}
UniPAD takes LiDAR point clouds $\mathcal{P}$ or multi-view images $\mathcal{I}$ as input. 
The input is first masked out by the mask generator (detailed in the following) and the visible parts are then fed into the modal-specific encoder.
For the point cloud $\mathcal{P}$, a point encoder, e.g., VoxelNet~\cite{yan2018second}, is adopted to extract hierarchical features $F_p$, as shown in Figure~\ref{fig:overall_stru}(a).
For images, features $F_c$ are extracted from $\mathcal{I}$ with a classic convolutional network, as illustrated in Figure~\ref{fig:overall_stru}(b).
To capture both high-level information and fine-grained details in data, we employ additional modality-specific FPN~\cite{lin2017fpn} to efficiently aggregate multi-scale features in practice.

\vspace{-0.8em}
\paragraph{Mask Generator}
\label{para:mask_generator}
Prior self-supervised approaches, as exemplified by He et al.~\cite{he2022mae}, have demonstrated that strategically increasing training difficulty can enhance model representation and generalization. Motivated by this, we introduce a mask generator as a means of data augmentation, selectively removing portions of the input.
Given points $\mathcal{P}$ or images $\mathcal{I}$, we adopt block-wise masking~\cite{yang2023gd-mae} to obscure certain regions.
Specifically, we first generate the mask according to the size of the output feature map, which is subsequently upsampled to the original input resolution.
For points, the visible areas are obtained by removing the information within the masked regions.
For images, we replace the traditional convolution with the sparse convolution as in \cite{tian2023spark}, which only computes at visible places.
After the encoder, masked regions are padded with zeros and combined with visible features to form regular dense feature maps.

\begin{figure*}[!t]
	\centering
	\includegraphics[width=1.65\columnwidth]{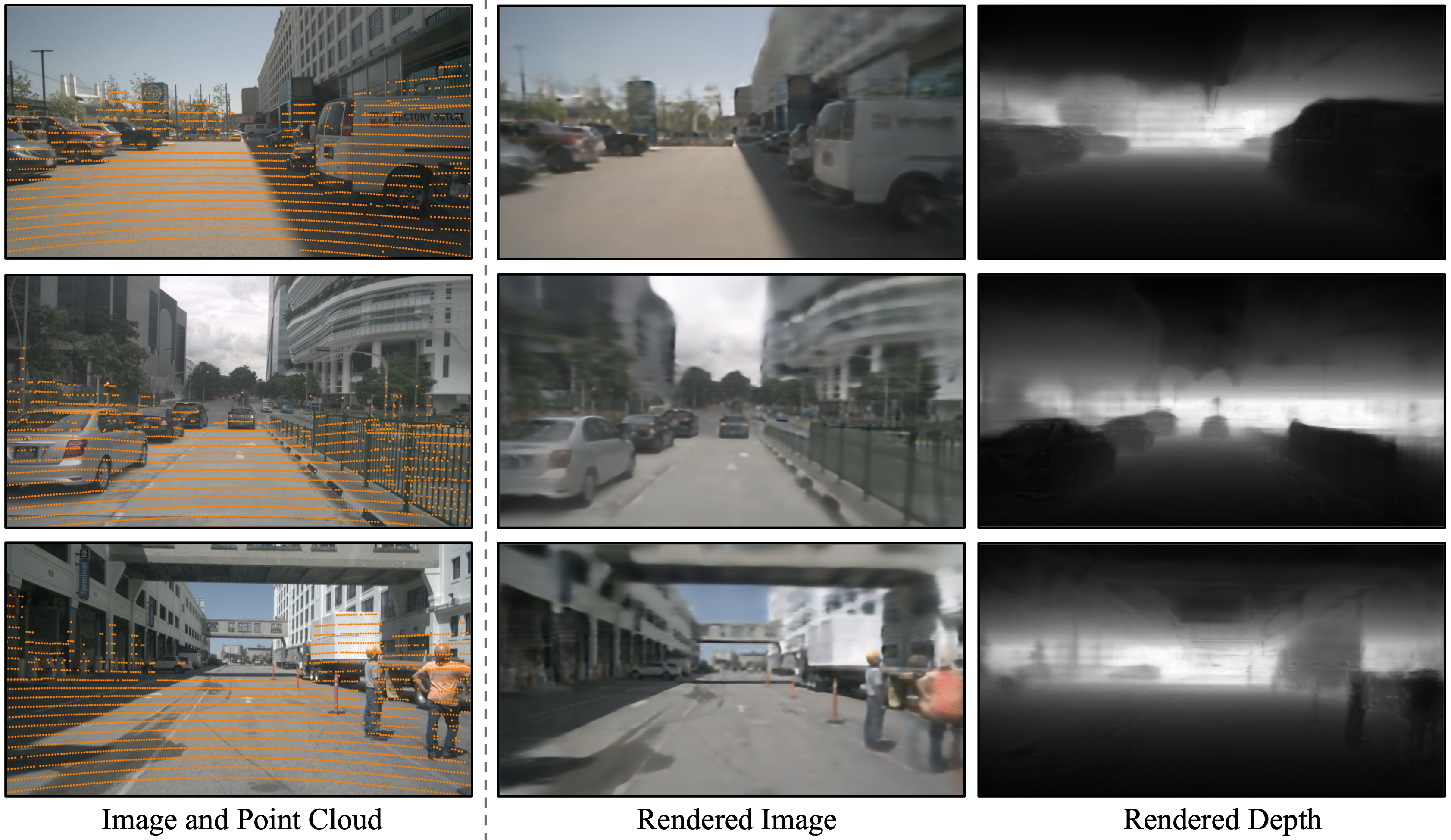}
	\vspace{-1.0em}
	\caption{Illustration of the rendering results, where the ground truth RGB and projected point clouds, rendered RGB, and rendered depth are shown on the left, middle, and right, respectively. }
	\label{fig:vis_result}
	\vspace{-1.6em}
\end{figure*}

\subsection{Unified 3D Volumetric Representation}
\label{subsec:unified_voxel_representation}
To make the pre-training method suitable for various modalities, it is crucial to find a unified representation. Transposing 3D points into the image plane would result in a loss of depth information, whereas merging them into the bird's eye view would lead to the omission of height-related details.
In this paper, we propose to convert both modalities into the 3D volumetric space, as shown in Figure~\ref{fig:overall_stru}(c), preserving as much of the original information from their corresponding views as possible.
For multi-view images, the view transformation~\cite{phi2020lss} is adopted to transform 2D features into the 3D ego-car coordinate system to obtain the volume features.
Specifically, we first predefine the 3D voxel coordinates $X_p\in\mathbb{R}^{X \times Y \times Z \times 3}$, where $X \times Y \times Z$ is the voxel resolution.
Subsequently, the $X_p$ is projected on multi-view images to index the corresponding 2D features, which are then multiplied by a learnable scaling factor.
The process can be calculated by:

 \begin{equation}
 	X'_p = T_{\mathrm{c2i}} T_{\mathrm{l2c}} X_p,\quad \mathcal{V}=\mathcal{B}(X'_p, F_c)\mathcal{T}(X'_p, \phi(F_c)),
\end{equation}
where $X'_p$ is the projected coordinates in the image plane, and $T_\mathrm{l2c}$ and $T_{\mathrm{c2i}}$ denote the transformation matrices from the LiDAR coordinate system to the camera frame and from the camera frame to image coordinates, respectively.
$\mathcal{V}$ is the constructed volumetric feature, $F_c$ is the image features, and $\phi$ is determined by a convolutional layer with a Softmax function.
$\mathcal{B}$ and $\mathcal{T}$ represent the bilinear and trilinear interpolation to retrieve the corresponding 2D features and scaling factor, respectively.
For the 3D point modality, we follow \cite{li2022uvtr} to directly retain the height dimension in the point encoder.
Finally, we leverage a projection layer involving $L$ conv-layers to enhance the voxel representation.

\subsection{Neural Rendering Decoder}
\label{subsec:volume_neural_rendering}

\paragraph{Differentiable Rendering}
We represent a novel use of neural rendering to flexibly incorporate geometry or textural clues into learned voxel features with a unified pre-training architecture, as shown in Figure~\ref{fig:overall_stru}(c).
Specifically, when provided the volumetric features, we sample some rays $\{\textbf{r}_i\}_{i=1}^K$ from multi-view images or point clouds and use differentiable volume rendering to render the color or depth for each ray.
The flexibility further facilitates the incorporation of 3D priors into the acquired image features, achieved via supplementary depth rendering supervision. This capability ensures effortless integration into both 2D and 3D frameworks.
Figure~\ref{fig:vis_result} shows the rendered RGB images and depth images based on our rendering decoder.

Inspired by \cite{wang2021neus}, we represent a scene as an implicit signed distance function (SDF) field to be capable of representing high-quality geometry details.
The SDF symbolizes the 3D distance between a query point and the nearest surface, thereby implicitly portraying the 3D geometry.
For ray $\textbf{r}_i$ with camera origin $\textbf{o}$ and viewing direction $\textbf{d}_i$, we sample $D$ ray points $\{\textbf{p}_{j} = \textbf{o} + t_{j} \textbf{d}_i \mid j=1,...,D, t_{j}<t_{j+1} \}$, where $\textbf{p}_{j}$ is the 3D coordinates of sampled points, and $t_j$ is the corresponding depth along the ray.
For each ray point $\textbf{p}_j$, the feature embedding $\textbf{f}_j$ can be extracted from the volumetric representation by trilinear interpolation.
Then, the SDF value $s_j$ is predicted by $\phi_\mathrm{SDF}(\textbf{p}_j, \textbf{f}_j)$, where $\phi_\mathrm{SDF}$ represents a shallow MLP.
For the color value, we follow ~\cite{oechsle2021unisurf} to condition the color field on the surface normal $\textbf{n}_j$ (i.e., the gradient of the SDF value at ray point $\textbf{p}_j$) and a geometry feature vector $\textbf{h}_i$ from $\phi_\mathrm{SDF}$.
Thus, the color representation is denoted as $c_j=\phi_\mathrm{RGB}(\textbf{p}_j, \textbf{f}_j, \textbf{d}_i, \textbf{n}_j, \textbf{h}_j)$, where $\phi_\mathrm{RGB}$ is parameterized by a MLP.
Finally, we render RGB value $\hat{Y}_i^\mathrm{RGB}$ and depth $\hat{Y}_i^\mathrm{depth}$ by integrating predicted colors and sampled depth along rays:
\begin{equation}
\label{eq:render}
	\hat{Y}_i^\mathrm{RGB}=\sum_{j=1}^D{w_j c_j}, \quad  \hat{Y}_i^\mathrm{depth}=\sum_{j=1}^D{w_j t_j},
\end{equation}
where $w_j$ represents an unbiased and occlusion-aware weight~\cite{wang2021neus} given by $w_j=T_j\alpha_j$.
$T_j=\prod_{k=1}^{j-1}(1-\alpha_k)$ is the accumulated transmittance, and $\alpha_j$ is the opacity value computed by:
\begin{equation}
	\alpha_j=\max \left(\frac{\sigma_s\left(s_j\right)-\sigma_s\left(s_{j+1}\right)}{\sigma_s\left(s_j\right)}, 0\right),
\end{equation}
where $\sigma_s(x)=(1+e^{-sx})^{-1}$ is a Sigmoid function modulated by a learnable parameter $s$.

\vspace{-0.8em}
\paragraph{Memory-friendly Ray Sampling}

Previous novel view synthesis methods prioritize dense supervision to enhance image quality. However, rendering a complete set of $S \times H \times W$ rays — where $S$ represents the number of camera views and $H \times W$ is the image resolution — presents substantial computational challenges, especially in the context of autonomous driving scenes.

\begin{figure}[!t]
    \centering
    \includegraphics[width=0.95\columnwidth]{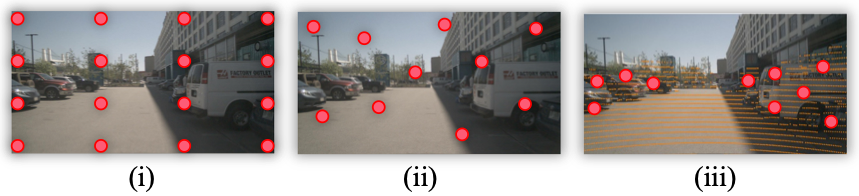}
    \vspace{-0.8em}
    \caption{Illustration of ray sampling strategies: i) dilation, ii) random, and iii) depth-aware sampling.}
    \label{fig:ray_sampling}
    \vspace{-1.2em}
\end{figure}

To alleviate computational challenges, we devise three memory-friendly ray sampling strategies to render a reduced subset of rays: \textit{Dilation Sampling}, \textit{Random Sampling}, and \textit{Depth-aware Sampling}, illustrated in Figure~\ref{fig:ray_sampling}.
1) \textit{Dilation Sampling} traverses the image at intervals of $I$, thereby reducing the ray count to $\frac{S \times H \times W}{I^2}$.
2) In contrast, \textit{Random Sampling} selects $K$ rays indiscriminately from all available pixels.
3) Although both dilation and random sampling are straightforward and significantly cut computation, they overlook the subtle prior information that is inherent to the 3D environment.
For example, instances on the road generally contain more relevant information over distant backgrounds like sky and buildings.
Therefore, we introduce \textit{depth-aware sampling} to selectively sample rays informed by available LiDAR information, bypassing the need for a full pixel set.
To implement this, we project point clouds onto the multi-view images and acquire the set of projection pixels with a depth less than the $\tau$ threshold.
Subsequently, rays are selectively sampled from this refined pixel set as opposed to the entire array of image pixels. 
In doing so, our approach not only alleviates the computational burden but also enhances the learned representation by concentrating on the most relevant segments within the scene.

\vspace{-0.8em}
\paragraph{Pre-training Loss}
The overall pre-training loss consists of the color loss and depth loss:
\begin{equation}
\begin{split}
	L&=\frac{\lambda_\mathrm{RGB}}{K}\sum_{i=1}^K|\hat{Y}_i^\mathrm{RGB}-Y_i^\mathrm{RGB}| \\
         &+ \frac{\lambda_\mathrm{depth}}{K^+}\sum_{i=1}^{K^+} |\hat{Y}_i^\mathrm{depth}-Y_i^\mathrm{depth}|,
\end{split}
\end{equation}
where $Y_i^\mathrm{RGB}$ and $Y_i^\mathrm{depth}$ are the ground-truth color and depth for each ray, respectively. $\hat{Y}_i^\mathrm{RGB}$ and $\hat{Y}_i^\mathrm{depth}$ are the corresponding rendered ones in Eq.~\ref{eq:render}.
 $K^+$ is the count of rays with available depth.

\section{Experiments}

\begin{table*}[t]
	\centering
	\caption{
		Comparisons of different methods with a single model on the nuScenes {\em val} set.
		We compare with classic methods on different modalities {\em without} test-time augmentation.
		$\dagger$: denotes our reproduced results based on  MMDetection3D~\cite{mmdet3d2020}.
		L, C, CS, and M indicate the LiDAR, Camera, Camera Sweep, and Multi-modality input, respectively.
	}
	\vspace{-1.0em}
        \tablestyle{10pt}{1.05}
        \begin{tabular}{l|c|c|c|c|cc}
		\toprule
		Methods & Present at & Modality & CS & CBGS & NDS$\uparrow$ & mAP$\uparrow$ \\
		\midrule
		PVT-SSD~\cite{yang2023pvtssd} & CVPR'23 & L & & \checkmark & 65.0 & 53.6 \\
		CenterPoint~\cite{yin2021center} & CVPR'21 & L & & \checkmark & 66.8 & 59.6 \\
		FSD~\cite{fan2022fsd} & NeurIPS'22 & L & &  \checkmark & 68.7 & 62.5 \\
		VoxelNeXt~\cite{chen2023voxelnext} & CVPR'23 & L & & \checkmark & 68.7 & 63.5 \\
		LargeKernel3D~\cite{chen2023largekernel3d} & CVPR'23 & L & & \checkmark & 69.1 & 63.3 \\
		TransFusion-L~\cite{bai2022transfusion} & CVPR'22 & L & & \checkmark & 70.1 & 65.1 \\
            CMT-L~\cite{yan2023cmt} & ICCV'23 & L & & \checkmark & 68.6 & 62.1 \\
		\rowcolor[gray]{.96}
		UVTR-L~\cite{li2022uvtr} & NeurIPS'22 & L & & \checkmark & 67.7 & 60.9 \\
		\rowcolor[gray]{.92}
		\textbf{UVTR-L+UniPAD (Ours)} & - & L & & \checkmark & \textbf{70.6} & \textbf{65.0} \\
		\midrule
		BEVFormer-S~\cite{li2022bevformer} & ECCV'22 & C & & \checkmark & 44.8 & 37.5 \\
		SpatialDETR~\cite{doll2022spatialdetr} & ECCV'22 & C & & & 42.5 & 35.1 \\
		PETR~\cite{liu2022petr} & ECCV'22 & C & & \checkmark & 44.2 & 37.0 \\
		Ego3RT~\cite{lu2022ego3rt} & ECCV'22 & C & & & 45.0 & 37.5 \\
		3DPPE~\cite{shu20233dppe} & ICCV'23 & C & & \checkmark & 45.8 & 39.1 \\
            BEVFormerV2~\cite{yang2023bevformerv2} & CVPR'23 & C & & & 46.7 & 39.6 \\
		CMT-C~\cite{yan2023cmt} & ICCV'23 & C & & \checkmark & 46.0 & 40.6 \\
		\rowcolor[gray]{.96}
		FCOS3D$^\dag$~\cite{wang2021fcos3d} & ICCVW'21 & C & & & 38.4 & 31.1 \\
		\rowcolor[gray]{.92}
		\textbf{FCOS3D+UniPAD (Ours)} & - & C & & & \textbf{40.1} & \textbf{33.2} \\
		\rowcolor[gray]{.96}
		UVTR-C~\cite{li2022uvtr} & NeurIPS'22  & C & & & 45.0 & 37.2 \\
		\rowcolor[gray]{.92}
		\textbf{UVTR-C+UniPAD (Ours)} & - & C & & & \textbf{47.4} & \textbf{41.5} \\
            \midrule
		\rowcolor[gray]{.96}
		UVTR-CS~\cite{li2022uvtr} & NeurIPS'22 & C & \checkmark & & 48.8 & 39.2 \\
		\rowcolor[gray]{.92}
		\textbf{UVTR-CS+UniPAD (Ours)} & - & C & \checkmark & & \textbf{50.2} & \textbf{42.8} \\
		\midrule
		PointPainting~\cite{vora2020pointpainting} & CVPR'20 & C+L & & \checkmark & 69.6 & 65.8 \\
		MVP~\cite{yin2021mvp} & NeurIPS'21 & C+L & & \checkmark & 70.8 & 67.1 \\
		TransFusion~\cite{bai2022transfusion} & CVPR'22 & C+L & & \checkmark & 71.3 & 67.5\\
		AutoAlignV2~\cite{chen2022autoalignv2} & ECCV'22 & C+L & & \checkmark & 71.2 & 67.1 \\
		BEVFusion~\cite{liang2022bevfusion} & NeurIPS'22 & C+L & & \checkmark & 71.0 & 67.9 \\
		BEVFusion~\cite{liu2023bevfusion} & ICRA'23 & C+L & & \checkmark & 71.4 & 68.5 \\
            ObjectFusion~\cite{cai2023objectfusion} & ICCV'23 & C+L & & \checkmark & 72.3 & 69.8 \\		
  DeepInteraction~\cite{yang2022deepinteraction} & NeurIPS'22 & C+L & & \checkmark & 72.6 & 69.9 \\
            SparseFusion~\cite{xie2023sparsefusion} & ICCV'23 & C+L & & \checkmark & 72.8 & 70.4 \\
		CMT-M~\cite{yan2023cmt} & ICCV'23 & C+L & & \checkmark & 72.9 & 70.3 \\
		\rowcolor[gray]{.96}
		UVTR-M~\cite{li2022uvtr} & NeurIPS'22 & C+L & & \checkmark & 70.2 & 65.4 \\
		\rowcolor[gray]{.92}
		\textbf{UVTR-M+UniPAD (Ours)} & - & C+L & & \checkmark & \textbf{73.2} & \textbf{69.9} \\
		\bottomrule
	\end{tabular}
	\label{tab:nuscene_val}
 \vspace{-2.2em}
\end{table*}

\begin{table}
\centering
\caption{Comparisons of different methods with a single model on the nuScenes segmentation dataset.}
\vspace{-1.0em}
\tablestyle{4.5pt}{1.05}
\begin{tabular}{l|c|c|cc}
    \toprule
     \multirow{2}{*}{Methods} & \multirow{2}{*}{Modality} & \multirow{2}{*}{Backbone} & \multicolumn{2}{c}{Split}  \\
     & & & \textit{val} & \textit{test} \\
    \midrule
    RangeFormer~\cite{kong2023rangformer} & L & Transformer & 78.1 & 80.1 \\
    SphereFormer~\cite{lai2023sphereformer} & L & Transformer & 78.4 & 81.9 \\
    WaffleIron~\cite{puy2023waffleiron} & L & Conv2D & 79.1 & - \\
    SPVNAS~\cite{tang2020spvnas} & L & SpConv & - & 77.4 \\
    Cylinder3D~\cite{zhu2021cylinder3d} & L & SpConv & 76.1 & 77.2 \\
    \rowcolor[gray]{.96}
    SpUNet~\cite{ChoyGS2019mink} & L & SpConv & 73.3 & - \\
    \rowcolor[gray]{.92}
    \textbf{SpUNet+UniPAD (Ours)} & L & SpConv & \textbf{79.4} & \textbf{81.1} \\
    \bottomrule
\end{tabular}
\label{tab:semseg}
\vspace{-2.8em}
\end{table}

\subsection{Datasets and Evaluation Metrics}

We conduct the experiments on the NuScenes~\cite{caesar2020nuscenes} dataset, which is a challenging dataset for autonomous driving.
It consists of 700 scenes for training, 150 scenes for validation, and 150 scenes for testing.
Each scene is captured through six different cameras, providing images with surrounding views, and is accompanied by a point cloud from LiDAR.
The dataset comes with diverse annotations, supporting tasks like 3D object detection and 3D semantic segmentation. 
For detection evaluation, we employ nuScenes detection score (NDS) and mean average precision (mAP), and for segmentation assessment, we use mean intersection-over-union (mIoU).

\subsection{Implementation Details}
We base our code on the MMDetection3D~\cite{mmdet3d2020} toolkit and train all models on 4 NVIDIA A100 GPUs. The input image is configured to $1600 \times 900$ pixels, while the voxel dimensions for point cloud voxelization are $[0.075, 0.075, 0.2]$.
During the pre-training phase, we implemented several data augmentation strategies, such as random scaling and rotation. Additionally, we partially mask the inputs, focusing only on visible regions for feature extraction.
The masking size and ratio for images are configured to $32$ and $0.3$, and for points to $8$ and $0.8$, respectively.
ConvNeXt-small~\cite{liu2022convnext} and VoxelNet~\cite{yan2018second} are adopted as the default image and point encoders, respectively.
A uniform voxel representation with the shape of $180 \times 180 \times 5$ is constructed across modalities. The feature projection layer reduces the voxel feature dimensions to $32$ via a $3$-kernel size convolution.
For the decoders, we utilize a $6$-layer MLP for SDF and a $4$-layer MLP for RGB. In the rendering phase, $512$ rays per image view and $96$ points per ray are randomly selected. We maintain the loss scale factors for $\lambda_\mathrm{RGB}$ and $\lambda_\mathrm{depth}$ at $10$.
The model undergoes training for $12$ epochs using the AdamW optimizer with initial learning rates of $2e^{-5}$ and $1e^{-4}$ for point and image modalities, respectively.
In the ablation studies, unless explicitly stated, fine-tuning is conducted for $12$ epochs on 50\% of the image data and for $20$ epochs on 20\% of the point data, without the use of CBGS~\cite{zhu2019cbgs} strategy and cut-and-paste~\cite{yan2018second} augmentation.

\subsection{Comparison with State-of-the-Art Methods}
\paragraph{3D Object Detection.}
In Table~\ref{tab:nuscene_val}, we compare UniPAD with previous detection approaches on the nuScenes validation set.
We adopt UVTR~\cite{li2022uvtr} as our baselines for point-modality (UVTR-L), camera-modality (UVTR-C), Camera-Sweep-modality (UVTR-CS), and fusion-modality (UVTR-M).
Benefits from the effective pre-training, UniPAD consistently improves the baselines, namely, UVTR-L, UVTR-C, and UVTR-M, by 2.9, 2.4, and 3.0 NDS, respectively.
When taking multi-frame cameras as inputs, UniPAD-CS brings 1.4 NDS and 3.6 mAP gains over UVTR-CS.
Our pre-training technique also achieves 1.7 NDS and 2.1 mAP improvements over the monocular-based baseline FCOS3D~\cite{wang2021fcos3d}.
Without any test time augmentation or model ensemble, our single-modal and multi-modal methods, UniPAD-L, UniPAD-C, and UniPAD-M, achieve impressive NDS of 70.6, 47.4, and 73.2, respectively, reaching the level of existing state-of-the-art methods.

\vspace{-0.8em}
\paragraph{3D Semantic Segmentation.}
In Table~\ref{tab:semseg}, we compare UniPAD with previous point cloud semantic segmentation approaches on the nuScenes Lidar-Seg dataset.
We adopt SpUNet~\cite{ChoyGS2019mink} implemented by Pointcept~\cite{pointcept2023} as our baseline.
Benefiting from effective pre-training, UniPAD improves the baselines by 6.1 mIoU, achieving state-of-the-art performance on the validation set. Meanwhile, UniPAD achieves an impressive mIoU of 81.1 on the \textit{test} set, which is comparable with existing state-of-the-art methods.

\subsection{Comparisons with Pre-training Methods.}
\paragraph{Image-based Pre-training.}
In Table~\ref{tab:image_pre_train}, we conduct comparisons between UniPAD and several other image-based pre-training approaches:
1) Depth Estimator: we follow \cite{park2021dd3d} to inject 3D priors into 2D learned features via depth estimation;
2) Detector: the image encoder is initialized using pre-trained weights from MaskRCNN~\cite{he2017maskrcnn} on the nuImages dataset~\cite{caesar2020nuscenes};
3) 3D Detector: the weights from the widely used monocular 3D detector~\cite{wang2021fcos3d} is used for model initialization, which relies on 3D labels for supervision.
UniPAD demonstrates superior knowledge transfer capabilities compared to previous unsupervised or supervised pre-training methods, showcasing the efficacy of our rendering-based pretext task.

\vspace{-0.8em}
\paragraph{Point-based Pre-training.}
For point modality, we also present comparisons with recently proposed self-supervised methods in Table~\ref{tab:point_pre_train}:
1) Occupancy-based: we implement ALSO~\cite{boulch2023also} in our framework to train the point encoder;
2) MAE-based: the leading-performing method~\cite{yang2023gd-mae} is adopted, which reconstructs masked point clouds using the chamfer distance.
3) Contrast-based: \cite{liu2021ppkt} is used for comparisons, which employs pixel-to-point contrastive learning to integrate 2D knowledge into 3D points.
Among these methods, UniPAD achieves the best NDS performance.
While UniPAD has a slightly lower mAP compared to the contrast-based method, it avoids the need for complex positive-negative sample assignments in contrastive learning.
More implementation details will be provided in the supplementary material.

\begin{table}
\centering
\caption{Comparison with different image-based pre-training.}
\vspace{-1.0em}
\tablestyle{9pt}{1.05}
\begin{tabular}{l|cc|cc}
    \toprule
    \multirow{2}{*}{Methods} & \multicolumn{2}{c|}{Label} & \multirow{2}{*}{NDS} & \multirow{2}{*}{mAP} \\
        & 2D & 3D & & \\
    \midrule
        UVTR-C (Baseline)  & & & 25.2 & 23.0 \\
        +Depth Estimator & & & 26.9\up{1.7} & 25.1\up{2.1} \\
        +Detector & \checkmark & & 29.4\up{4.2} & 27.7\up{4.7} \\
        +3D Detector & & \checkmark & 31.7\up{6.5} & 29.0\up{6.0} \\
        \midrule
    \textbf{+UniPAD} & & & 32.9\up{7.7} & 32.6\up{9.6} \\
    \bottomrule 
\end{tabular}
\label{tab:image_pre_train}
\vspace{-1.2em}
\end{table}

\begin{table}
\centering
\caption{Comparison with different point-based pre-training.}
\vspace{-1.0em}
\tablestyle{9pt}{1.05}
\begin{tabular}{l|cc|cc}
    \toprule
    \multirow{2}{*}{Methods} & \multicolumn{2}{c|}{Support} & \multirow{2}{*}{NDS} & \multirow{2}{*}{mAP} \\
        & 2D & 3D & & \\
    \midrule
    UVTR-L (Baseline) & & & 46.7 & 39.0 \\
    +Occupancy-based & & \checkmark & 48.2\up{1.5} & 41.2\up{2.2} \\
    +MAE-based & & \checkmark & 48.8\up{2.1} & 42.6\up{3.6} \\
    +Contrast-based & \checkmark & \checkmark & 49.2\up{2.5} & 48.8\up{9.8} \\
    \midrule
    \textbf{+UniPAD} & \checkmark & \checkmark & 55.8\up{9.1} & 48.1\up{9.1} \\
    \bottomrule 
\end{tabular}
\label{tab:point_pre_train}
\vspace{-2.2em}
\end{table}

\subsection{Effectiveness on Various Backbones.}

\paragraph{Different View Transformations.}
In Table~\ref{tab:view_trans}, we investigate different view transformation strategies for converting 2D features into 3D space, including BEVDet~\cite{huang2021bevdet}, BEVDepth~\cite{li2023bevdepth}, and BEVformer~\cite{li2022bevformer}.
Due to the prevalent use of BEV representation, we integrate these methods into our framework by transforming features into volumetric representations.
Consistent improvements ranging from 5.2 to 6.3 NDS can be observed across different transformation techniques, which demonstrates the strong generalization ability of the proposed approach.

\vspace{-0.8em}
\paragraph{Different Modalities.}
Unlike most previous pre-training methods, our framework can be seamlessly applied to various modalities.
To verify the effectiveness of our approach, we set UVTR as our baseline, which contains detectors with point, camera, and fusion modalities.
Table~\ref{tab:diff_modality} shows the impact of UniPAD on different modalities.
UniPAD consistently improves the UVTR-L, UVTR-C, and UVTR-M by 9.1, 7.7, and 6.9 NDS, respectively.

\vspace{-0.8em}
\paragraph{Scaling up Backbones.}
To test UniPAD across different backbone scales, we adopt an off-the-shelf model, ConvNeXt, and its variants with different numbers of learnable parameters.
As shown in Table~\ref{tab:scale_up_back}, one can observe that with our UniPAD pre-training, all baselines are improved by large margins of
+6.0$\sim$7.7 NDS and +8.2$\sim$10.3 mAP.
The steady gains suggest that UniPAD has the potential to boost various state-of-the-art networks.

\begin{table}
\centering
\caption{Pre-training effect on different view transformations.}
\vspace{-1.0em}
\tablestyle{10pt}{1.05}
\begin{tabular}{l|c|cc}
    \toprule
    Methods & View Transform & NDS & mAP \\
    \midrule
        BEVDet & Pooling & 27.1 & 24.6 \\
        \textbf{+UniPAD} & Pooling & 32.7\up{5.6} & 32.8\up{8.2} \\
        \midrule
        BEVDepth & Pooling \& Depth & 28.9 & 28.1 \\
        \textbf{+UniPAD} & Pooling \& Depth & 34.1\up{5.2} & 33.9\up{5.8} \\
        \midrule
        BEVformer & Transformer & 26.8 & 24.5  \\
        \textbf{+UniPAD} & Transformer & 33.1\up{6.3} & 31.9\up{7.4} \\
    \bottomrule 
\end{tabular}
\label{tab:view_trans}
\vspace{-1.2em}
\end{table}

\begin{table}
\centering
\caption{Pre-training effectiveness on different input modalities.}
\vspace{-1.0em}
\tablestyle{11pt}{1.05}
\begin{tabular}{l|c|cc}
    \toprule
    Methods & Modality & NDS & mAP \\
    \midrule
    UVTR-L & LiDAR & 46.7 & 39.0 \\
    \textbf{+UniPAD} & LiDAR & 55.8\up{9.1} & 48.1\up{9.1} \\
    \midrule
    UVTR-C & Camera & 25.2 & 23.0 \\
    \textbf{+UniPAD} & Camera & 32.9\up{7.7} & 32.6\up{9.6} \\
    \midrule
    UVTR-M & LiDAR-Camera & 49.9 & 52.7 \\
    \textbf{+UniPAD} & LiDAR-Camera & 56.8\up{6.9} & 57.0\up{4.3} \\
    \bottomrule 
\end{tabular}
\label{tab:diff_modality}
\vspace{-2.2em}
\end{table}

\begin{table}[!b]
        \vspace{-1.8em}
	\centering
	\caption{Pre-training effectiveness on different backbone scales.}
	\vspace{-1.0em}
        \tablestyle{2.2pt}{1.05}
        \begin{tabular}{l|ccc}
			\toprule
			\multirow{2}{*}{Methods} & \multicolumn{3}{c}{Backbone} \\
                & ConvNeXt-S & ConvNeXt-B & ConvNeXt-L \\
                \midrule
                UVTR-C & 25.2/23.0 & 26.9/24.4 & 29.1/27.7 \\
                \textbf{+UniPAD} & 32.9\up{7.7}/32.6\up{9.6} & 34.1\up{7.2}/34.7\up{10.3} & 35.1\up{6.0}/35.9\up{8.2} \\
			\bottomrule 
	\end{tabular}
	\label{tab:scale_up_back}
\end{table}

\subsection{Ablation Studies}

\begin{table}[!t]
	\centering
	\caption{Ablation studies of the masking ratio.}
	\vspace{-1.0em}
        \tablestyle{12pt}{1.05}
        \begin{tabular}{l|ccccc}
			\toprule
			Ratio & 0.1 & 0.3 & 0.5 & 0.7 & 0.9 \\
                \midrule
                NDS & 31.9 & \textbf{32.9} & 32.3 & 32.1 & 31.4 \\
			\bottomrule 
	\end{tabular}
	\label{tab:mask_ratio}
 \vspace{-1.2em}
\end{table}

\begin{table}[!t]
	\centering
	\caption{Ablation studies of the decoder depth.}
	\vspace{-1.0em}
        \tablestyle{10.3pt}{1.05}
        \begin{tabular}{l|ccccc}
			\toprule
			Layer & (2, 2) & (4, 3) & (5, 4) & (6, 4) & (7, 5) \\
                \midrule
                NDS & 31.3 & 31.9 & 32.1 & \textbf{32.9} & 32.7 \\
			\bottomrule 
	\end{tabular}
	\label{tab:decoder_depth}
 \vspace{-1.2em}
\end{table}

\begin{table}[!t]
	\centering
	\caption{Ablation studies of the decoder width.}
	\vspace{-1.0em}
        \tablestyle{12.2pt}{1.05}
        \begin{tabular}{l|ccccc}
			\toprule
			Dim. & 16 & 32 & 64 & 128 & 256 \\
                \midrule
                NDS & 32.1 & \textbf{32.9} & 32.5 & \textbf{32.9} & 32.4 \\
			\bottomrule 
	\end{tabular}
	\label{tab:decoder_width}
 \vspace{-2.0em}
\end{table}

\paragraph{Masking Ratio.}
Table~\ref{tab:mask_ratio} shows the influence of the masking ratio for the camera modality.
We discover that a masking ratio of 0.3, which is lower than the ratios used in previous MAE-based methods, is optimal for our method.
This discrepancy could be attributed to the challenge of rendering the original image from the volume representation, which is more complex compared to image-to-image reconstruction.
For the point modality, we adopt a mask ratio of 0.8, as suggested in \cite{yang2023gd-mae}, considering the spatial redundancy inherent in point clouds.

\vspace{-0.8em}
\paragraph{Rendering Design.}
Our examinations in Tables \ref{tab:decoder_depth}, \ref{tab:decoder_width}, and \ref{tab:rendering_tech} illustrate the flexible design of our differentiable rendering.
In Table~\ref{tab:decoder_depth}, we vary the depth $(D_\mathrm{SDF}, D_\mathrm{RGB})$ of the SDF and RGB decoders, revealing the importance of sufficient decoder depth for succeeding in downstream detection tasks.
This is because deeper ones may have the ability to adequately integrate geometry or appearance cues during pre-training.
Conversely, as reflected in Table~\ref{tab:decoder_width}, the width of the decoder has a relatively minimal impact on performance.
Thus, the default dimension is set to $32$ for efficiency.
Additionally, we explore the effect of various rendering techniques in Table~\ref{tab:rendering_tech}, which employ different ways for ray point sampling and accumulation.
Using NeuS~\cite{wang2021neus} for rendering records a 0.4 and 0.1 NDS improvement compared to UniSurf~\cite{oechsle2021unisurf} and VolSDF~\cite{lior2021volsdf}, respectively, showcasing the learned representation can be improved by utilizing well-designed rendering methods and benefiting from the advancements in neural rendering.

\begin{table}[!b]
\vspace{-1.4em}
	\centering
	\caption{Ablation studies of the rendering technique.}
	\vspace{-1.0em}
        \tablestyle{20pt}{1.05}
        \begin{tabular}{l|cc}
            \toprule
            Methods & NDS & mAP \\
            \midrule
            UniSurf~\cite{oechsle2021unisurf} & 32.5 & 32.1 \\
            VolSDF~\cite{lior2021volsdf} & 32.8 & 32.4 \\
            NeuS~\cite{wang2021neus} & \textbf{32.9} & \textbf{32.6} \\
            \bottomrule
        \end{tabular}
	\label{tab:rendering_tech}
\end{table}

\vspace{-0.8em}
\paragraph{Memory-friendly Ray Sampling.}
Instead of rendering the entire set of multi-view images, we sample only a subset of rays to provide supervision signals.
Table~\ref{tab:sampling_str} outlines the different strategies explored to minimize memory usage and computational costs during pre-training.
Our observations indicate that depth-aware sampling holds a substantial advantage, improving scores by 0.4 and 1.0 NDS compared to random sampling ($K=512$) and dilation sampling ($I=16$), respectively.
The sampling excludes regions without well-defined depth, like the sky, from contributing to the loss.
This allows the representation learning to focus more on the objects in the scene, which is beneficial for downstream tasks.
Meanwhile, it costs less memory usage than dilation sampling.

\vspace{-0.8em}
\paragraph{Feature Projection.}
The significance of feature projection is shown in Table~\ref{tab:feat_proj}.
Removing projection from pre-training and fine-tuning leads to drops of 1.8 and 2.7 NDS, respectively, underscoring the essential role it plays in enhancing voxel representation.
Concurrently, utilizing shared parameters for the projection during pre-training and fine-tuning induces reductions of 0.8 NDS and 0.6 mAP.
This phenomenon is likely due to the disparity between the rendering and recognition tasks, with the final layers being more tailored for extracting features specific to each task.

\begin{table}[!t]
	\centering
	\caption{Ablation studies of the sampling strategy.}
	\vspace{-1.0em}
        \tablestyle{7.8pt}{1.05}
        \begin{tabular}{l|ccc}
            \toprule
            Methods & Memory &NDS & mAP \\
            \midrule
            Dilation Sampling & 1.4$\times$ & 31.9 & 32.4 \\
            Random Sampling & \textbf{1$\times$} & 32.5 & 32.1 \\
            Depth-aware Sampling & \textbf{1$\times$} & \textbf{32.9} & \textbf{32.6} \\
            \bottomrule
        \end{tabular}
	\label{tab:sampling_str}
\vspace{-1.2em}
\end{table}

\begin{table}[!t]
	\centering
	\caption{Ablation studies of the feature projection.}
	\vspace{-1.0em}
        \tablestyle{14pt}{1.05}
        \begin{tabular}{l|cc}
            \toprule
            Methods & NDS & mAP \\
            \midrule
            Baseline & \textbf{32.9} & \textbf{32.6} \\
            w/o Projection$_{\mathrm{FT}}$ & 30.2\down{2.7} & 29.7\down{2.9} \\
            w/o Projection$_\mathrm{PT}$ & 31.1\down{1.8} & 30.5\down{2.1} \\
            Shared Projection & 32.1\down{0.8} & 32.0\down{0.6} \\
            \bottomrule
        \end{tabular}
	\label{tab:feat_proj}
\vspace{-1.2em}
\end{table}

\begin{table}[!t]
	\centering
	\caption{Ablation studies of the pre-trained components.}
	\vspace{-1.0em}
        \tablestyle{11pt}{1.05}
        \begin{tabular}{l|cc}
            \toprule
    	Methods & NDS & mAP \\
    	\midrule
            Baseline  & 25.2  & 23.0\\
            +Encoder & 32.0\up{6.8} & 31.8\up{8.8} \\
            +Encoder \& FPN   & 32.2\up{0.2}  & 32.2\up{0.4}  \\
    	+Encoder \& FPN \& VT & \textbf{32.9}\up{0.7} & \textbf{32.6}\up{0.4} \\
            \bottomrule
        \end{tabular}
	\label{tab:pretrained_com}
\vspace{-2.em}
\end{table}

\vspace{-0.8em}
\paragraph{Pre-trained Components.}
In Table~\ref{tab:pretrained_com}, the influence of pre-trained parameters on each component is investigated. Replacing the pre-trained weights of the FPN and view transformation (VT) with those from a random initialization induces declines of 0.2 and 0.7 NDS, respectively, thereby highlighting the crucial roles of these components.

\section{Conclusion}
We present UniPAD, an innovative self-supervised learning paradigm that excels in various 3D perception tasks.
UniPAD stands out for its ingenious adaptation of NeRF as a unified rendering decoder, enabling seamless integration into both 2D and 3D frameworks. 
The inherent adaptability of our approach bridges the 2D and 3D domains, which could facilitate representation learning through advancements in the other domain.
For instance, semantic knowledge can be infused into point clouds via additional semantic supervision, leveraging the outputs of well-developed models like SAM~\cite{kirillov2023sam} in the 2D domain as learning targets.

\vspace{-0.8em}
\paragraph{Limitation.}
There are still certain limitations to the approach.
For instance, we need to explicitly transform point and image features into volumetric representations, which would increase memory usage as voxel resolution rises.
\vspace{-6pt}
\paragraph{Acknowledgement}
This work was supported in part by The National Nature Science Foundation of China (Grant Nos: 62273303), in part by Yongjiang Talent Introduction Programme (Grant No: 2022A-240-G), in part by Ningbo Key R\&D Program (No. 2023Z231, 2023Z229), in part by the National Key R\&D Program of China (NO. 2022ZD0160101).

{
    \small
    \bibliographystyle{ieeenat_fullname}
    \bibliography{main}
}



\end{document}